# Hybrid Bayesian Neural Networks with Functional Probabilistic Layers

Daniel T. Chang (张遵)

*IBM (Retired)* dtchang43@gmail.com

**Abstract:** Bayesian neural networks provide a direct and natural way to extend standard deep neural networks to support probabilistic deep learning through the use of probabilistic layers that, traditionally, encode weight (and bias) uncertainty. In particular, hybrid Bayesian neural networks utilize standard deterministic layers together with few probabilistic layers judicially positioned in the networks for uncertainty estimation. A major aspect and benefit of Bayesian inference is that priors, in principle, provide the means to encode prior knowledge for use in inference and prediction. However, it is difficult to specify priors on weights since the weights have no intuitive interpretation. Further, the relationships of priors on weights to the functions computed by networks are difficult to characterize. In contrast, functions are intuitive to interpret and are direct since they map inputs to outputs. Therefore, it is natural to specify priors on functions to encode prior knowledge, and to use them in inference and prediction based on functions. To support this, we propose hybrid Bayesian neural networks with functional probabilistic layers that encode function (and activation) uncertainty. We discuss their foundations in functional Bayesian inference, functional variational inference, sparse Gaussian processes, and sparse variational Gaussian processes. We further perform few proof-of-concept experiments using GPflus, a new library that provides Gaussian process layers and supports their use with deterministic Keras layers to form hybrid neural network and Gaussian process models.

## 1 Introduction and Overview

*Bayesian neural networks* [1] provide a direct and natural way to extend standard deep neural networks to support *probabilistic deep learning* [2] through the use of *probabilistic layers* that, traditionally, encode weight (and bias) uncertainty. For example, using Keras and TensorFlow Probability (TFP), as shown in [1], we can replace a deterministic Keras layer (e.g. Convolution2D) with a corresponding probabilistic TFP layer (e.g. Convolution2DFlipout).

The main problem with Bayesian neural networks is that the architecture of deep neural networks makes it quite redundant, and costly, to account for uncertainty for a large number of successive layers. *Hybrid Bayesian neural networks* [1], which utilize standard deterministic layers together with few probabilistic layers judicially positioned in the networks, provide a practical solution. For example, using Keras and TFP, we can simply add a probabilistic TFP layer (e.g. DenseVariational) to the end of a deterministic Keras neural network to generate *predictions with uncertainty estimates*.

A major aspect and benefit of Bayesian inference is that *priors*, in principle, provide the means to encode prior knowledge for use in inference and prediction [11]. However, it is difficult to specify *priors on weights* since the weights have no intuitive interpretation. Further, the relationships of priors on weights to the functions computed by networks are difficult to characterize.

In contrast, functions are intuitive to interpret and are direct since they map inputs to outputs. Therefore, it is natural to specify *priors on functions* to encode prior knowledge. Existing approaches to functional priors [11] focus on developing them for use in inference and prediction based on weights. Unfortunately, the relationships of functional priors to the weights of networks are difficult to characterize.

To support the use of functions to specify priors, which encode prior knowledge, and to use functions directly in inference and prediction, we propose *hybrid Bayesian neural networks with functional probabilistic layers* that encode *function uncertainty* (and *activation uncertainty* [3]).

We discuss their foundations in functional Bayesian inference, functional variational inference, sparse Gaussian processes, and sparse variational Gaussian processes. The formulation of the *functional Bayesian inference* and the *functional variational inference* are obtained from the Bayesian inference and the variational inference [1], respectively, by replacing the stochastic parameters **θ** with the stochastic function f [6].

*Gaussian processes*, being generalization of Gaussian distributions, are the natural choice for use as functional priors. However, *Gaussian process inference*, using *Gaussian process priors*, has two main issues [4]. First, it lacks a closed form functional posteriors when the functional likelihood is non-Gaussian. Second, it suffers from a cubic time complexity $\mathcal{O}(N^3)$ [5], where N being the number of data points.

*Sparse Gaussian processes* [6, 4-5] are based on partitioning the infinite set of random variables represented by a Gaussian process into two sets. The first set contains a *finite subset* evaluated at a finite set of *evaluation points* of size M, with M << N. The second set contains the *remaining infinite set* evaluated at all points except for the finite evaluation points.

*Sparse variational Gaussian processes* [6] are a framework for *functional variational inference* based on sparse Gaussian processes. They are applicable to any functional likelihood and they are scalable for large data sets, reducing the time complexity from $\mathcal{O}(N^3)$ to $\mathcal{O}(M^3)$.

We perform few proof-of-concept experiments using *GPflus* [7-8], a new library that provides Gaussian process layers and supports their use with deterministic Keras layers to form *hybrid neural network and Gaussian process models*. GPflux relies on *GPflow* [9-10] for most of its Gaussian process objects and operations. GPflow supports Sparse variational Gaussian processes.

# 2 Hybrid Bayesian Neural Networks (HBNNs)

The main problem with Bayesian neural networks is that the architecture of deep neural networks makes it quite redundant, and costly, to account for uncertainty for a large number of successive layers. Hybrid Bayesian neural networks (HBNNs) [1], which utilize standard deterministic layers together with few probabilistic layers judicially positioned in the networks, provide a practical solution.

## 2.1 HBNNs with Weight Uncertainty

In the simplest architecture, *HBNNs with weight uncertainty* [1] (or *hBNNs* for short) can be represented as follow, with **x** as input variable, **y** as target or output variable, and **h** as hidden variables:

- $\mathbf{h}_0 = \mathbf{x}$
- $\mathbf{h}_i = f^a_i(\mathbf{W}_i \mathbf{h}_{i-1} + \mathbf{b}_i) \quad \forall i \in$ [deterministic layers]
- $\mathbf{h}_j = f^a_j(\mathbf{W}_j \mathbf{h}_{j-1} + \mathbf{b}_j) \quad \forall j \in$ [probabilistic layers, at least one], $\mathbf{W}_j \sim p(\mathbf{W}_j)$, $\mathbf{b}_j \sim p(\mathbf{b}_j)$; $j \neq i$, $(i, j) \in [1, L]$
- $\mathbf{y} = \mathbf{h}_L$

where (at each layer) **h** is represented as a linear transformation of the previous one, followed by a nonlinear *activation (function)* $f^a$. The *weights* **W** and *biases* **b**, which are denoted by the *parameters* **θ**, of probabilistic layers are *stochastic*.

Probabilistic layers with weight uncertainty are trained using *Bayesian inference*, in particular *variational inference*, as discussed in [1]. A major problem with weight uncertainty is that it is difficult to specify *priors on weights*. First, the weights have no intuitive interpretation. Second, the relationships of priors on weights to the functions computed by networks are difficult to characterize.

## 2.2 HBNNs with Functional Probabilistic Layers

*HBNNs with functional probabilistic layers* (or *hfBNNs* for short) can be represented as:

- $\mathbf{h}_0 = \mathbf{x}$
- $\mathbf{h}_i = f^a_i(\mathbf{W}_i \mathbf{h}_{i-1} + \mathbf{b}_i) \quad \forall i \in$ [deterministic layers]
- $\mathbf{h}_j = f_j(\mathbf{h}_{j-1}) \quad \forall j \in$ [functional probabilistic layers, at least one], $f_j \sim p(f_j)$; $j \neq i$, $(i, j) \in [1, L]$
- $\mathbf{y} = \mathbf{h}_L$

The *function f* of a functional probabilistic layer is *stochastic*.

In addition to *function uncertainty*, functional probabilistic layers can be used to encode *activation uncertainty* [3]. A probabilistic layer that encodes activation uncertainty:

- $\mathbf{h}_j = f^a_j(\mathbf{W}_j \mathbf{h}_{j-1} + \mathbf{b}_j)$   $f^a_j \sim p(f^a_j)$

is a special case of a functional probabilistic layer, and can be represented by a linear deterministic layer (with no activation) followed by a functional probabilistic layer using the activation function:

- $\mathbf{l}_j = \mathbf{W}_j \mathbf{h}_{j-1} + \mathbf{b}_j$   linear deterministic layer
- $\mathbf{h}_j = f^a_j(\mathbf{l}_j)$   $f^a_j \sim p(f^a_j)$   functional probabilistic layer using activation function

Functional probabilistic layers are trained using *functional Bayesian inference*, in particular *functional variational inference*, as discussed in the next section.

## 3 Functional Variational Inference

For simplicity and generality, we denote a *functional probabilistic layer*, $\mathbf{h}_j = f_j(\mathbf{h}_{j-1})$ $f_j \sim p(f_j)$, as

$$\mathbf{y} = f(\mathbf{x}) \quad f \sim p(f)$$

with $\mathbf{x}$ representing the input variable and $\mathbf{y}$ the output variable. We note that each layer has its *own functional prior* $p(f)$ and *own inference* for f. All f's are related to each other, and to deterministic layers, only through hidden variables $\mathbf{h}$'s, which are inputs / outputs.

The formulation of the *functional Bayesian inference* and the *functional variational inference* are obtained from the Bayesian inference and the variational inference [1], respectively, by replacing the stochastic parameters $\boldsymbol{\theta}$ with the stochastic function f, as is similarly done in [6].

### 3.1 Functional Bayesian Inference

The *functional Bayesian inference* is based on the stochastic *function* f. The main idea is to infer a *posterior distribution (functional posterior)* over the function f given some observed data D using Bayes theorem as:

$$p(f \mid D) = p(D \mid f)p(f) / p(D) = p(D \mid f)p(f) / \int p(D \mid f)p(f)df,$$

where $p(D \mid f)$ is the *functional likelihood*, $p(f)$ is the *functional prior*, and $p(D)$ is the *marginal likelihood* (*evidence*).

The functional posterior can be used to model new unseen data $D^*$ using the *functional posterior predictive*:

$$p(D^* \mid D) = \int p(D^* \mid f)p(f \mid D)df,$$

which is also called the *Bayesian model average* because it averages the predictions of all plausible models weighted by their posterior probability.

For supervised learning, the *functional posterior* can be written as:

$$p(f \mid D) = p(D_y \mid D_x, f)p(f) / \int p(D_y \mid D_x, f)p(f)df,$$

where $D_x$ is the training features, $D_y$ the training labels, and $p(D_y \mid D_x, f)$ is the *functional likelihood*. The *functional posterior predictive* can be computed as:

$$p(y \mid x, D) = \int p(y \mid x, f)p(f \mid D)df$$

where $p(y \mid x, f)$ is the *functional predictive*.

## 3.2 Functional Variational Inference

Computing the functional posterior $p(f \mid D)$ is generally intractable. Usually, *functional variational inference*, an approximate functional Bayesian inference, is used. It learns a *functional variational distribution* $q_\phi(f)$, parameterized by a set of parameters $\phi$, to approximate the functional posterior $p(f \mid D)$. The values of the parameters $\phi$ is inferred such that the functional variational distribution is as close as possible to the functional posterior. The measure of closeness used is the *functional KL-divergence*:

$$D_{KL}(q_\phi(f) \parallel p(f \mid D)).$$

Minimizing this is equivalent to maximizing the *functional ELBO*:

$$ELBO = \log(p(D)) - D_{KL}(q_\phi(f) \parallel p(f \mid D)).$$

The functional KL-divergence, between two stochastic processes of infinite dimension, is difficult to work with in general. Usually, *sparse variational Gaussian processes* are employed to obtain tractable approximate solutions, as discussed in the next section.

## 4 Sparse Variational Gaussian Processes

*Sparse variational Gaussian processes* are a framework for *functional variational inference* based on *sparse Gaussian processes*. They are applicable to any functional likelihood and they are scalable for large data sets.

### 4.1 Gaussian Processes

The *Gaussian process (GP)* is a well-known *nonparametric* probabilistic model. A Gaussian process is a generalization of the *Gaussian distribution*. Whereas a probability distribution describes random variables which are scalars or vectors, a *stochastic process* (e.g., Gaussian process) governs the properties of *functions*. We can loosely think of a function as *an infinite vector*, each entry in the vector specifying the function value f($x_i$) at a particular input $x_i$. A key aspect of this is that if we ask only for the properties of the function at *a finite number of points*, then *Gaussian process inference* will give us the same answer if we ignore the infinitely many other points.

In the *function-space view* a Gaussian process defines a *distribution over functions*, and inference takes place directly in the *space of functions*. A Gaussian process is completely specified by its *mean function* (average of all functions) and its symmetric, positive-definite *kernel (covariance) function* (how much individual functions can vary around the mean function):

$$\mu(\mathbf{x}) = \mathbb{E}[f(\mathbf{x})],$$

$$k(\mathbf{x}, \mathbf{x'}) = \mathbb{E}[(f(\mathbf{x}) - \mu(\mathbf{x}))(f(\mathbf{x'}) - \mu(\mathbf{x'}))] = \text{cov}(f(\mathbf{x}), f(\mathbf{x'}))$$

As such, we write the Gaussian process as

$$f(\mathbf{x}) \sim \mathcal{GP}(\mu(\mathbf{x}), k(\mathbf{x}, \mathbf{x'})),$$

and denote the corresponding marginal distribution over f as p(f).

It is important to note that if we evaluate the Gaussian process at any finite subset $\{X_1, X_2, ..., X_M\}$ of the input domain $\mathcal{X}$ with cardinality M, we would obtain an *M-dimensional multivariate Gaussian random variable* **f** [5-6]:

$$\mathbf{f} \sim \mathcal{N}(\boldsymbol{\mu}_\mathbf{f}, \mathbf{K}_\mathbf{ff}),$$

where $\boldsymbol{\mu}_\mathbf{f}$ and $\mathbf{K}_\mathbf{ff}$ are the *mean vector* and *covariance matrix* obtained by evaluating the mean and covariance function respectively at $\{X_1, X_2, ..., X_M\}$, i.e. $\boldsymbol{\mu}_\mathbf{f}[i] = \mu(X_i)$ and $\mathbf{K}_\mathbf{ff}[i, j] = k(X_i, X_j)$. We denote the corresponding marginal distribution over **f** as p(**f**).

Gaussian processes, being generalization of Gaussian distributions, are the natural choice for use as functional priors. However, *Gaussian process inference*, using *Gaussian process priors*, has two main issues [4]. First, it lacks a closed form functional posteriors when the functional likelihood is non-Gaussian. Second, it suffers from a cubic time complexity $\mathcal{O}(N^3)$ [5], where N being the number of data points, because of the inversion and determinant of the N × N kernel matrix $\mathbf{K}_\mathbf{NN}$. Both issues can be addressed within the framework of *variational inference* using *sparse Gaussian processes*.

## 4.2 Sparse Gaussian Processes

Sparse Gaussian processes [6, 4-5] are based on partitioning the infinite set of random variables represented by a Gaussian process into two sets. The first set contains a *finite subset* denoted as **u** evaluated at a finite set of *evaluation points* $\{Z_1, Z_2, ..., Z_M\} \in \mathcal{X}$, such that $\mathbf{u}[m] = f(Z_m)$, with mean vector $\boldsymbol{\mu}_\mathbf{u}$ and covariance matrix $\mathbf{K}_\mathbf{uu}$. The second set contains the *remaining infinite set* denoted as f(.) evaluated at all points in $\mathcal{X}$ except for the finite points $Z_m$.

Given marginal distributions p(**u**) with mean vector $\boldsymbol{\mu}_\mathbf{u}$ and covariance matrix $\mathbf{K}_\mathbf{uu}$, and p(f(.)) with mean function $\mu(.)$ and covariance function $k(., \cdot)$, the *conditional Gaussian process* of f(.) conditioned on **u** can be obtained as [6]:

$$f(.)|\mathbf{u} \sim \mathcal{GP}(\mu(.) + \mathbf{k}_{.\mathbf{u}}\mathbf{K}_\mathbf{uu}^{-1}(\mathbf{u} - \boldsymbol{\mu}_\mathbf{u}), k(., \cdot) - \mathbf{k}_{.\mathbf{u}}\mathbf{K}_\mathbf{uu}^{-1}\mathbf{k}_{\mathbf{u}.}),$$

where $\mathbf{k}_{.\mathbf{u}}$ and $\mathbf{k}_{\mathbf{u}.}$ denote vector-valued functions that express the cross-covariance between **u** and f(.), i.e. $\mathbf{k}_{.\mathbf{u}}[m] = k(., Z_m)$ and $\mathbf{k}_{\mathbf{u}.}[m] = k(Z_m, .)$. We denote the corresponding conditional distribution as p(f(.)|**u**).

Assuming another marginal distribution q(**u**), e.g. an approximate marginal distribution, with mean vector $\mathbf{m}_\mathbf{u}$ and covariance matrix $\mathbf{S}_\mathbf{uu}$, we can integrate p(f(.)|**u**) over **u** with q(**u**):

$$q(f(.)) = \int p(f(.)|\mathbf{u})q(\mathbf{u})d\mathbf{u},$$

resulting in the standard definition of a *sparse Gaussian process* [6]:

$$f(.) \sim \mathcal{GP}(\mu(.) + \mathbf{k}_{,u}\mathbf{K}_{uu}^{-1}(\mathbf{m}_u - \boldsymbol{\mu}_u), k(., .\cdot) - \mathbf{k}_{,u}\mathbf{K}_{uu}^{-1}(\mathbf{K}_{uu} - \mathbf{S}_{uu})\mathbf{K}_{uu}^{-1}\mathbf{k}_{u,\cdot}).$$

The evaluation points $Z_m$ are usually called "*inducing points*", and $\mathbf{u}$ is called "*inducing variables*". The number M of inducing points $Z_m$ governs the expressiveness of the sparse Gaussian process and the computational costs. More inducing points mean more pseudo-training examples and hence a more accurate approximate representation. However, since M determines the dimension of $\mathbf{u}$, more inducing points, on the other hand, mean higher memory requirements and higher computational complexity $\mathcal{O}(M^3)$. Practical considerations favor $M << N$, with N being the number of training data.

In practice, q(f(.)) is used to approximate an intractable functional posterior p(f(.) | D) through *variational inference*. Variational inference treats an approximate inference problem as an optimization problem where *inducing points* $Z_m$, as well as the *mean vector* $\mathbf{m}_u$ and the *covariance matrix* $\mathbf{S}_{uu}$ of the inducing variable distribution q($\mathbf{u}$), are treated as optimization arguments that are automatically determined in the course of training. This is discussed next.

## 4.3 Sparse Variational Gaussian Processes (SVGP)

For supervised learning and given training data, $D = \{y_n, X_n\}_{n=1}^{N}$, the *functional ELBO for sparse Gaussian processes* can be written as [6]:

$$\text{ELBO}(\phi) = \log(p(D)) - D_{KL}(q_\phi(f) \parallel p(f | D))$$

$$= \sum_{n=1}^{N} \int q_\phi(f(.)) \ln p(y_n | f(.), X_n) \, df(.) - D_{KL}(q_\phi(f(.)) \parallel p(f(.))),$$

where the variational parameters $\phi$ refer to *inducing points* $Z_m$ as well as the *mean vector* $\mathbf{m}_u$ and the *covariance matrix* $\mathbf{S}_{uu}$ of the inducing variable distribution q($\mathbf{u}$).

Since $y_n$ is conditioned on f(.) evaluated at $X_n$, and does not depend on function values at evaluation points other than $X_n$, the first term can be written as [6]:

$$\sum_{n=1}^{N} \int q_\phi(f(X_n)) \ln p(y_n | f(X_n)) \, df(X_n).$$

The second KL-divergence term represents the KL divergence between f(.) under the approximate posterior $q_\phi(f(.))$ and under the prior p(f(.)), and is mathematically equivalent to the finite-dimensional KL divergence between the variational distribution $q_\phi(\mathbf{u})$ over the inducing variables $\mathbf{u}$ and the distribution over $\mathbf{u}$ under the prior process $p(\mathbf{u})$. Thus the *SVGP ELBO* can be written as [6]:

$$\text{ELBO}(\phi) = \Sigma_{n=1}^{N} \int q_\phi(f(X_n)) \ln p(y_n \mid f(X_n))\, df(X_n) - D_{KL}(q_\phi(\mathbf{u}) \parallel p(\mathbf{u})).$$

A practical advantage of the SVGP ELBO is that the KL-divergence term has an analytical expression because of the Gaussian assumptions.

*Predictions* $\{y^*_n\}_{n=1}^{N^*}$ for a new data set $\{X^*_n\}_{n=1}^{N^*}$ are straightforward by multiplying the likelihood with the approximate posterior Gaussian process and integrating over function values [6]:

$$p(y^*_1, \ldots, y^*_{N^*} \mid X^*_1, \ldots, X^*_{N^*}) = \Sigma_{n=1}^{N^*} \int q_\phi(\mathbf{f}^*) \ln p(y^*_n \mid f(X^*_n))\, d\mathbf{f}^*,$$

where $q_\phi(\mathbf{f}^*)$ denotes the multivariate Gaussian obtained when evaluating the approximate posterior Gaussian process $q_\phi(f(.))$ at the new observations $X^*_1, \ldots, X^*_{N^*}$.

## 5 Experiments

We perform few proof-of-concept experiments to qualitatively compare standard deep neural networks (*DNNs*), hybrid Bayesian neural networks with weight uncertainty (*hBNNs*), and hybrid Bayesian neural networks with functional probabilistic layers (*hfBNNs*), using Keras, TensorFlow Probability (TFP), and GPflux, respectively. GPflux [7-8] is a new library supporting SVGP and hybrid neural network and Gaussian process modeling, which we briefly overview first.

### 5.1 GPflux

GPflux relies on *GPflow* [9-10] for most of its Gaussian process objects and operations. GPflow contains well-tested and stable implementations of key Gaussian process building blocks, such as *mean functions, kernels*, and *inducing variables*. GPflow uses *variational inference* as the primary approximation method, and supports *sparse Gaussian processes* which ensure that the approximation is scalable and close, in a KL divergence sense, to the functional posterior.

GPflux is compatible with and built on top of *Keras*. It is designed as a *deep learning library* where functionality is packed into *layers*. The key building block in GPflux is the *GPLayer*, which represents the *prior and posterior* of a single (multi-output) Gaussian process. It can be seen as the analogue of a standard fully-connected (dense) layer in a standard neural network, but with an infinite number of basis functions. It is defined by a *MeanFunction*, *Kernel*, and *InducingVariables*, which are all GPflow objects.

A GPLayer's call takes the output of the previous layer, $\mathbf{h}_{j-1}$, but returns an object that represents the *complete Gaussian distribution* of $\mathbf{h}_j = f_j(\mathbf{h}_{j-1})$. If a subsequent layer is not able to use the distributional output, a *sample* will be taken using the reparametrization trick. This functionality is provided by TensorFlow Probability's *DistributionLambda* layer.

GPflux also provides other *Bayesian and Gaussian process layers*. It is possible to combine GPflux layers with standard neural network components in Keras, such as convolutional or fully-connected layers. In most cases one can directly make use of Keras' *Sequential* API to combine the different GPflux and Keras layers into a *hybrid model*.

Deep neural networks are trained by minimizing the prediction discrepancy for examples in a training set. This is a similar to the first term in the SVGP ELBO, which is passed to the GPplus framework using *LikelihoodLoss()*. The second KL-divergence term of the SVGP ELBO is added to the loss by the GPLayer calls.

## 5.2 Experiments

For experiments, we use the *"Hybrid Deep GP models: combining GP and Neural Network layers"* tutorial code provided in GPflix [8] as the base. We modify the code as appropriate for each experiment.

*DNNs*

The DNN example consists of three Keras Dense layers:

```
model = tf.keras.Sequential(
    [
        tf.keras.layers.Dense(100, activation="relu"),
        tf.keras.layers.Dense(100, activation="relu"),
        tf.keras.layers.Dense(1),
    ]
)
loss = tf.keras.losses.MeanSquaredError()

model.compile(loss=loss, optimizer="adam")
hist = model.fit(X, Y, epochs=500, verbose=0)
```

The predictions (point estimates) by the model are shown below. It can be seen that the *prediction accuracy is very good*, being able to predict the major pattern in the training data.

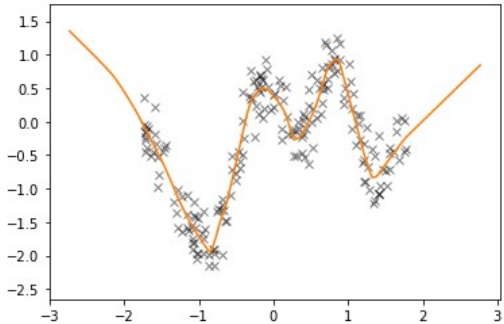

*hBNNs*

For the first example, we replace the last Keras Dense layer in the DNN model with a TFP DenseVariational layer with a Gaussian prior and a Gaussian approximate posterior (see [1] for the definitions of prior and posterior):

```
model = tf.keras.Sequential(
    [
        tf.keras.layers.Dense(100, activation="relu"),
        tf.keras.layers.Dense(100, activation="relu"),
        tfp.layers.DenseVariational(
            units=2,
            make_prior_fn=prior,
            make_posterior_fn=posterior,
            kl_weight=1 / num_data,
        ),
        tfp.layers.IndependentNormal(1)
    ]
)
def negative_loglikelihood(targets, estimated_distribution):
    return -estimated_distribution.log_prob(targets)
model.compile(
    optimizer=tf.keras.optimizers.RMSprop(learning_rate=0.001),
    loss=negative_loglikelihood,
    metrics=[tf.keras.metrics.RootMeanSquaredError()],
)
hist = model.fit(X, Y, epochs=500, verbose=0)
```

The predictions made by the model are shown below. It can be seen that the prediction accuracy (mean) is not as good as that of DNN. However, the *uncertainty estimates (95% confidence intervals) are very good*, covering the training data spread well.

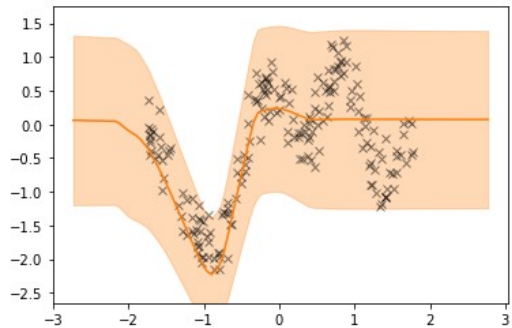

For the second example, we append a TFP DenseVariational layer to the last Keras Dense layer in the DNN model:

```
model = tf.keras.Sequential(
    [
        tf.keras.layers.Dense(100, activation="relu"),
        tf.keras.layers.Dense(100, activation="relu"),
        tf.keras.layers.Dense(1),
        tfp.layers.DenseVariational(
            units=2,
            make_prior_fn=prior,
            make_posterior_fn=posterior,
            kl_weight=1 / num_data,
        ),
        tfp.layers.IndependentNormal(1)
    ]
)
```

The predictions made by the model are shown below. It can be seen that the *prediction accuracy is better* than the first example, and the *uncertainty estimates are also better*.

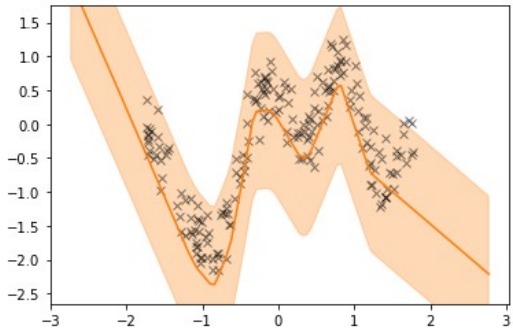

### hfBNNs

For all hfBNN examples, we append a GPflux GPLayer to the last Keras Dense layer in the DNN model. We are not able to replace the last Keras Dense layer with a GPLayer as it raises an error.

The first example is identical to the one used in the GPflux tutorial:

```
kernel = gpflow.kernels.SquaredExponential()
inducing_variable = gpflow.inducing_variables.InducingPoints(
    np.linspace(X.min(), X.max(), num_inducing).reshape(-1, 1)
)

likelihood = gpflow.likelihoods.Gaussian(0.001)
likelihood_container = gpflux.layers.TrackableLayer()
likelihood_container.likelihood = likelihood

model = tf.keras.Sequential(
    [
        tf.keras.layers.Dense(100, activation="relu"),
        tf.keras.layers.Dense(100, activation="relu"),
        tf.keras.layers.Dense(1, activation="linear"),
        gpflux.layers.GPLayer(
            kernel, inducing_variable, num_data=num_data,
            num_latent_gps=output_dim
        ),
        likelihood_container,
    ]
)
loss = gpflux.losses.LikelihoodLoss(likelihood)

model.compile(loss=loss, optimizer="adam")
hist = model.fit(X, Y, epochs=500, verbose=0)
```

The predictions made by the model are shown below. It can be seen that the *prediction accuracy is very good*, nearly the same as the DNN model, but the *uncertainty estimates are very poor except for out-of-bound data*, not being able to cover the training data spread at all.

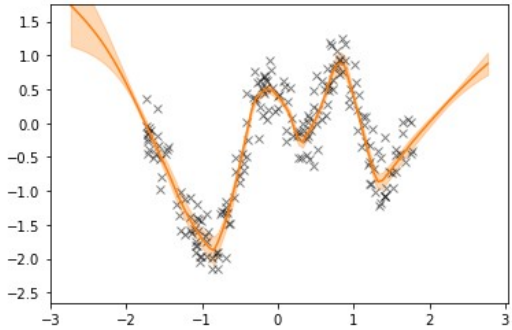

For the second example, we add a second GPLayer, effectively using deep Gaussian processes:

```
model = tf.keras.Sequential(
    [
        tf.keras.layers.Dense(100, activation="relu"),
        tf.keras.layers.Dense(100, activation="relu"),
        tf.keras.layers.Dense(1, activation="linear"),
        gpflux.layers.GPLayer(
            kernel, inducing_variable, num_data=num_data,
            num_latent_gps=output_dim
        ),
        gpflux.layers.GPLayer(
            kernel, inducing_variable, num_data=num_data,
            num_latent_gps=output_dim
        ),
        likelihood_container,
    ]
)
```

The predictions made by the model are shown below. It can be seen that they are nearly the same as the first example which uses a single GPLayer.

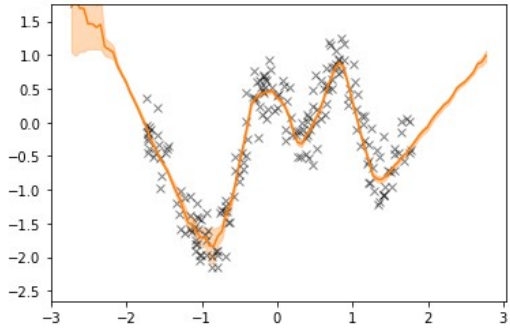

For the third example, as the first example, we use a single GPLayer but with an ArcCosine kernel:

```
kernel = gpflow.kernels.ArcCosine()
inducing_variable = gpflow.inducing_variables.InducingPoints(
    np.linspace(X.min(), X.max(), num_inducing).reshape(-1, 1)
)
```

The predictions made by the model are shown below. It can be seen that it has *better uncertainty estimates, though not good enough*, than the first example which uses a SquaredExponential kernel. (BTW, the predictions made by using a Polynomial kernel are nearly the same as that by using a SquaredExponential kernel .)

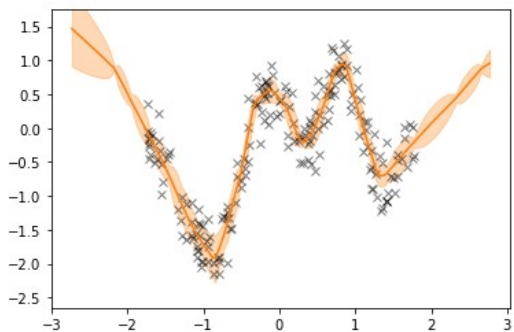

For the fourth example, as the first example, we use a single GPLayer but with InducingPatches as inducing variables:

```
kernel = gpflow.kernels.SquaredExponential()
inducing_variable = gpflow.inducing_variables.InducingPatches(
    np.linspace(X.min(), X.max(), num_inducing).reshape(-1, 1)
)
```

The predictions made by the model are shown below. It can be seen that they are nearly the same as the first example which uses InducingPoints as inducing variables.

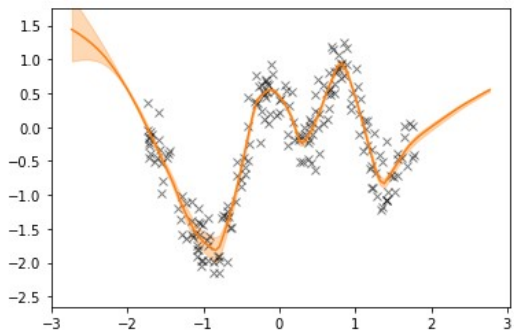

## 6 Summary and Conclusion

To support the use of functions to specify priors, which encode prior knowledge, and to use functions directly in inference and prediction, we propose hybrid Bayesian neural networks with functional probabilistic layers that encode function (and activation) uncertainty.

We discuss their foundations in functional Bayesian inference, functional variational inference, sparse Gaussian processes, and sparse variational Gaussian processes. We further perform few proof-of-concept experiments using GPflus, a new library that provides Gaussian process layers and supports their use with deterministic Keras layers to form hybrid neural network and Gaussian process models.

GPflux is a promising, and the first as far as we know, library for use in building hybrid Bayesian neural networks with functional probabilistic layers. However, GPflux is very new and, for hybrid neural network and Gaussian process modeling, it currently has only one use case. More use cases are needed to explore the hybrid modeling usage, capability and limitation.

Also, documentation is needed that elucidates the usage patterns and usage constraints of hybrid modeling in GPflux. Based on the findings of our proof-of-concept experiments, uncertainty estimation is a critical area that needs major improvement in GPflux.

**Acknowledgement:** Thanks to my wife Hedy (郑期芳) for her support.